\documentclass[10pt,twocolumn,letterpaper]{article}

\usepackage{cvpr}              %

\usepackage{graphicx}
\usepackage{amsmath}
\usepackage{amssymb}
\usepackage{booktabs}

\usepackage[pagebackref,breaklinks,colorlinks]{hyperref}

\usepackage[capitalize]{cleveref}
\crefname{section}{Sec.}{Secs.}
\Crefname{section}{Section}{Sections}
\Crefname{table}{Table}{Tables}
\crefname{table}{Tab.}{Tabs.}

\usepackage{xcolor}
\definecolor{bittersweet}{rgb}{1.0, 0.44, 0.37}
\definecolor{mygreen}{rgb}{0.29, 0.7, 0.48}

\usepackage{algorithm}
\usepackage{algpseudocode}
\usepackage{listings}
\usepackage{pifont}

\usepackage{tabularx}
\usepackage{multirow}
\usepackage{paralist}
\usepackage{makecell}
\usepackage{graphicx}
\usepackage{amsmath}
\usepackage{gensymb}
\usepackage{xspace}
\usepackage{pifont}%
\newcommand{\cmark}{\ding{51}}%
\newcommand{\xmark}{\ding{55}}%
\usepackage{booktabs} 
\usepackage{listings}
\usepackage{xcolor}

\def\ModelName{\textsc{MARVAL}\xspace}
\def\Dagger{DAGGER\xspace}
\def\Marky{Marky\xspace}

\newcommand{\pz}{\hphantom{.0}}

\newcommand{\secref}[1]{Section~\ref{#1}}
\newcommand{\figref}[1]{Figure~\ref{#1}}
\newcommand{\tabref}[1]{Table~\ref{#1}}

\newcommand{\A}{{\cal A}}

\newcommand{\W}{{\cal W}}

\def\cvprPaperID{12100} %
\def\confName{CVPR}
\def\confYear{2023}

\begin{document}

\title{A New Path: Scaling Vision-and-Language Navigation with\\Synthetic Instructions and Imitation Learning}

\author{
{\normalsize \textbf{Aishwarya Kamath$^{*\dagger1}$ \quad Peter Anderson$^{*2}$ \quad Su Wang$^2$ \quad Jing Yu Koh$^{\dagger3}$ \quad Alexander Ku$^2$}}\\
{\normalsize \textbf{Austin Waters$^2$ \quad Yinfei Yang$^{\dagger4}$ \quad Jason Baldridge$^2$ \quad Zarana Parekh$^2$}}\\
$^1$New York University \quad $^2$Google Research \quad $^3$Carnegie Mellon University \quad $^4$Apple
}

\maketitle

\begin{abstract}{\let\thefootnote\relax\footnote{$^*$Equal Contribution. $^\dagger$Work done while at Google Research.}}
Recent studies in Vision-and-Language Navigation (VLN) train RL agents to execute natural-language navigation instructions in photorealistic environments, as a step towards robots that can follow human instructions.
However, given the scarcity of human instruction data and limited diversity in the training environments, these agents still struggle with complex language grounding and spatial language understanding.
Pretraining on large text and image-text datasets from the web has been extensively explored but the improvements are limited. 
We investigate large-scale augmentation with synthetic instructions.
We take 500+ indoor environments captured in densely-sampled 360\degree{} panoramas, construct navigation trajectories through these panoramas, and generate a visually-grounded instruction for each trajectory using \Marky{}~\cite{marky2022}, a high-quality multilingual navigation instruction generator. We also synthesize image observations from novel viewpoints using an image-to-image GAN~\cite{koh2022simple}. The resulting dataset of 4.2M instruction-trajectory pairs is two orders of magnitude larger than existing human-annotated datasets, and contains a wider variety of environments and viewpoints. To efficiently leverage data at this scale, we train a simple transformer agent with imitation learning. On the challenging RxR dataset, our approach outperforms all existing RL agents, improving the state-of-the-art NDTW from 71.1 to 79.1 in seen environments, and from 64.6 to 66.8 in unseen test environments. Our work points to a new path to improving instruction-following agents, emphasizing large-scale training on near-human quality synthetic instructions.
\vspace{-5mm}
\end{abstract}

\section{Introduction}

\begin{figure}
  \centering
  \includegraphics[width=1\linewidth]{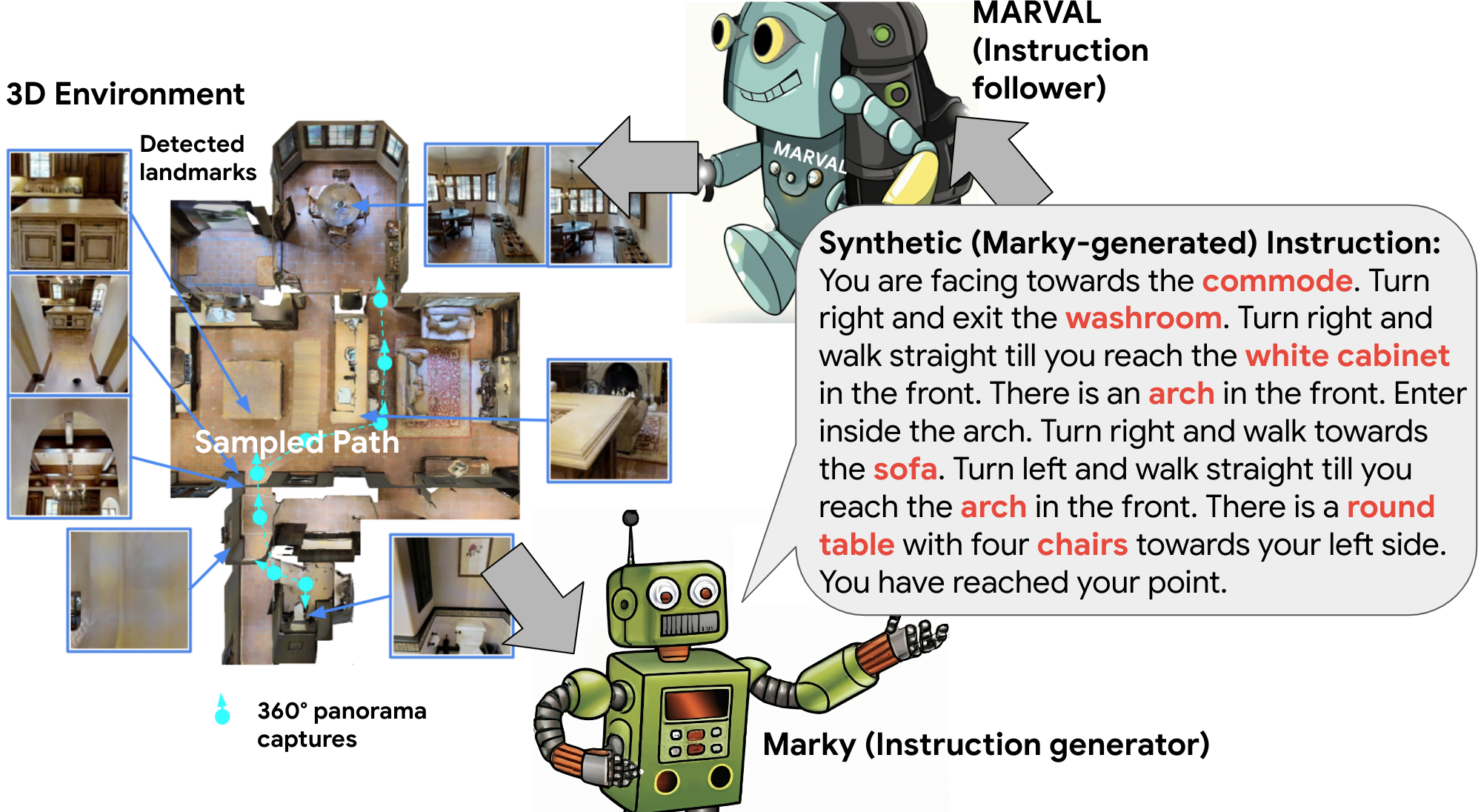}
  \caption{Simpler agents with more data: We investigate large-scale augmentation using 500+ environments annotated with synthetic instructions that approach human quality.}
  \label{fig:concept}
  \vspace{-5mm}
\end{figure}

Developing intelligent agents that follow human instructions is a long-term, formidable challenge in AI \cite{winograd1971procedures}. 
A recent focus addressing this problem space is Vision-and-Language Navigation (VLN) \cite{anderson_vision-and-language_2018, chen2019touchdown}. Navigation is an ideal test bed for studying instruction-following, since the task can be simulated photo-realistically at scale and evaluation is straightforward. However, datasets that capture the linguistic diversity and idiosyncrasies of real human instructors are small and expensive to collect.

Shortages of human-annotated training data for other vision-and-language tasks have been partially addressed by pretraining transformers on up to billions of image-text pairs. This has underpinned dramatic improvements in image captioning \cite{Zhang2021VinVLRV, Wang2021SimVLMSV}, visual question answering \cite{Tan2019LXMERTLC}, phrase grounding \cite{Kamath2021MDETRM, li2021grounded}, text-to-video retrieval, video question answering \cite{lei2021less} and text-to-image synthesis \cite{dalle,parti}. However, these are all static image or video tasks, whereas VLN agents \textit{interact} with 3D environments. In VLN, pretraining on large image-text and text-only datasets has been thoroughly explored \cite{majumdar2020improving, NEURIPS2020_56dc0997, Hong2021VLNBERTAR, moudgil2021soat}, but improvements are more limited. Arguably, progress in VLN has plateaued while still leaving a large gap between machine and human performance~\cite{zhu_diagnosing_2021}.
We hypothesize that static image-text and text-only datasets -- despite their size -- lack the spatially grounded and action-oriented language needed for effective VLN pretraining.
Consider instructions from the Room-across-Room (RxR) dataset \cite{AlexKu:20}, which illustrate that wayfinding requires an understanding of allocentric and egocentric spatial expressions (\textit{near a grey console table \textbf{behind you}}), verbs (\textit{\textbf{climb} the stairs}), imperatives and negations (\textit{\textbf{do not} enter the room in front}) and temporal conditions (\textit{walk \textbf{until} you see an entrance on your left}). Such expressions are rarely found in image-text datasets. Though similar expressions are found in text-only corpora, their meaning as it relates to the physical world is hard to infer from text alone (without sensorimotor context) \cite{Bisk2020}.

To address this problem, we investigate large-scale augmentation with synthetic in-domain data, i.e., model-generated navigation instructions for trajectories in realistic 3D environments using previously developed components \cite{koh2022simple, marky2022}. 
We construct a large dataset using \Marky{}~\cite{marky2022}, which generates VLN instructions that approach the quality of human instructors.
\cite{marky2022} released the 1M \Marky{} instruction-trajectory pairs situated in 61 Matterport3D \cite{Matterport3D} environments.  
To increase the diversity of the environments (and thus the scenes and objects available in them), 
we automatically annotate an additional 491 environments from the Gibson dataset \cite{xiazamirhe2018gibsonenv}. 
Gibson environments have been underutilized in prior VLN work due to the lack of navigation graphs indicating navigable trajectories through its densely-sampled 360\degree{} panoramas. We train a model that classifies navigable directions for Matterport3D and use it to construct the missing navigation graphs. We sample 3.2M trajectories from these graphs and annotate them with \Marky{}. To further increase the variability of trajectories, we synthesize image observations from novel viewpoints using an image-to-image GAN \cite{koh2022simple}. The resulting dataset is two orders of magnitude larger than existing human-annotated ones, and contains a wider variety of scenes and viewpoints. We have released our Gibson navigation graphs and the Marky-Gibson dataset.\footnote{\href{https://github.com/google-research-datasets/RxR/tree/main/marky-mT5}{//github.com/google-research-datasets/RxR/tree/main/marky-mT5}}

With orders of magnitude more training examples and environments, we explore VLN agent performance with imitation learning (IL), i.e., behavioral cloning and \Dagger{} \cite{ross2011reduction} %
IL can take advantage of high-throughput transformer frameworks such as T5 \cite{2020t5} and thus efficiently train on 4.2M instructions (accumulating over 700M steps of experience). This is a departure from most prior VLN work in low-data settings, e.g. \cite{chen2021hamt} report that pure IL underperforms by 8.5\% success rate compared to agents trained with both IL and online reinforcement learning (RL) algorithms such as A3C \cite{Mnih2016AsynchronousMF}. However, IL outperforms RL in related tasks with sufficient training data \cite{rramrakhya2022}. Online RL also requires interacting with the environment at each step; this precludes efficient data prefetching and parallel preprocessing and thus imposes unavoidable overhead compared to IL.
Empirically, we confirm that training existing models such as HAMT \cite{chen2021hamt} on 4.2M instructions is infeasible without ground-up re-engineering, though we do find incorporating 10K additional synthetic instructions into HAMT training modestly improves performance. 
Training with IL aligns with the trend towards large-scale multi-task vision-and-language models trained with supervised learning; these have unified tasks as diverse as visual question answering, image captioning, object detection, image classification, OCR and text reasoning \cite{pali} -- and could include VLN in future.

Experimentally, in detailed ablation studies we show that adding Gibson environments, synthesizing additional image observations from novel viewpoints, increasing the capacity of the transformer, and finetuning with \Dagger{} all improve agent performance. On the challenging RxR dataset -- which contains multilingual instructions with a median trajectory length of 15m -- our best agent \textit{using only imitation learning} outperforms all prior RL agents. Evaluating on novel instruction-trajectories in seen environments (Val-Seen), we improve over the state-of-the-art by 8\%, reaching 79.1 NDTW. In new, unseen environments (Test), we improve by 2\%, achieving 66.8 NDTW. We also show that that self-training with synthetic instructions in new environments (still without human annotations) improves performance by an additional 2\% to 68.6 NDTW. Overall, our RxR results point to a new path to improving instruction-following agents, emphasizing large-scale training on near-human quality synthetic instructions. Perhaps surprisingly, on the English-only R2R dataset \cite{anderson_vision-and-language_2018}, our IL agent achieves strong but not state-of-the-art results. Marky{} was trained on RxR, so we attribute this to domain differences between R2R and RxR, underscoring the domain dependence of synthetic instructions.

\section{Related Work}

\noindent\textbf{Vision-and-Language Navigation}\quad
Agents that follow instructions by navigating to a prescribed location were initially studied in simple settings requiring limited or no perception, using instructions that were often procedurally generated \cite{chen2011learning, Artzi2013WeaklySL, Andreas2015AlignmentBasedCS, Mei2016ListenAA, misra2017mapping}. More recent work has explored photorealistic 3D settings and natural language instructions \cite{anderson_vision-and-language_2018, chen2019touchdown}, using environments such as Matterport3D \cite{Matterport3D} and Streetview \cite{mehta2020retouchdown}. This instantiation of the problem, known as Vision-and-Language Navigation (VLN), raises the prospect of sim-to-real transfer to physical robots \cite{anderson2021sim}, and encouraged further datasets exploring dialog \cite{thomason:corl19,hahn2020way}, object search \cite{qi_reverie_2020} and multilingual instructions \cite{AlexKu:20}.

\noindent\textbf{Pretraining and Transfer Learning}\quad
The use of realistic imagery and language in VLN, combined with the cost of collecting human instruction annotations, leads to a natural focus on pretraining and transfer learning to improve performance. Majumdar et al.~\cite{majumdar2020improving} formulate VLN as an instruction-trajectory alignment problem, and initialize a transformer model using pretrained BERT weights~\cite{Devlin2019BERTPO} then perform additional pretraining on image-text pairs from Conceptual Captions~\cite{sharma-etal-2018-conceptual}. Li et al.~\cite{li2019robust} also use a BERT model, although more recent approaches have favored learned text encoders from scratch by pretraining with Masked Language Modeling (MLM) and related objectives on instruction-trajectory data \cite{hao_towards_2020, chen2021hamt}. In terms of image representations, early work \cite{fried2018speaker} used ResNet features \cite{he2015deep} pretrained on ImageNet \cite{ILSVRC15}, although pretrained object detectors have also been explored \cite{Hong2021VLNBERTAR, majumdar2020improving, li2022clear} (typically a Faster-RCNN \cite{Ren2015FasterRT}). More recently, Chen et al.~\cite{chen2021hamt} use a vision transformer (ViT) \cite{Dosovitskiy2021AnII} and current-state-of-the art agents \cite{Shen2021HowMC, Li2022EnvEditEE} use CLIP \cite{Radford2021LearningTV},
obtaining improvements over similarly sized encoders pretrained on ImageNet. However, although pretraining and transfer learning from large text and image-text datasets has been thoroughly explored, a significant gap to human performance remains.

\noindent\textbf{Data Augmentation}\quad
Fried et al.~\cite{fried2018speaker} were the first to demonstrate that performance following human instructions could be improved by augmenting training with synthetic (model-generated) instructions. A variety of other data augmentation approaches have been investigated, including modifying existing environments before generating new instructions~\cite{backtranslate2019,Li2022EnvEditEE}, training on a synthetic dataset of path-instruction pairs generated using online rental listings~\cite{Guhur2021AirbertIP}, and training with a generative model that infills and outpaints spatially perturbed panos of indoor environments to generate new observations~\cite{koh2022simple,koh2021pathdreamer} (we also use this in \secref{sec:experiments}). Notwithstanding these contributions, prior work incorporating synthetic instructions has been severely limited by instruction quality and scale. In human wayfinding evaluations, the instructions used \cite{fried2018speaker,backtranslate2019} were shown to be surprisingly weak, being poorly grounded and mostly unfollowable by people~\cite{Zhao21}. The recently proposed \Marky{} model~\cite{marky2022} (an instruction generator trained with text-aligned visual landmark correspondences) addresses this limitation, achieving near-human quality on R2R-style paths in unseen environments. We address the second limitation (scale) by developing an automated pipeline for scaling navigation graphs to 500+ new environments which we annotate with 3.2M instructions, and training agents on two orders of magnitude more data than before. Using this approach, we achieve state-of-the-art results in a VLN setting using a purely imitation learning agent. In contrast, most recent VLN work focuses on RL agents. An exception is DUET~\cite{chen2022think}, which uses imitation learning in conjunction with a global action space based on a topological map.

\section{Approach}
\label{sec:approach}

\noindent\textbf{Problem set up}\quad
The agent is instantiated in an environment and must follow a natural language instruction $\W$. At time step $t$, the agent receives observation $o_t$ and chooses action $a_t$ that transitions it from state $s_t$ to new state $s_{t+1}$. Following prior work, each observation is a photorealistic panoramic image (hereafter, pano) encoded as 36 image feature vectors $o_t{=}\{I^o_{t,1}, I^o_{t,2}, ... , I^o_{t,K}\}$. These features are extracted from perspective projections at 36 view angles (12 headings $\times$ 3 elevations at 30\degree{} intervals).
The agent moves by choosing an action $a_t$ from a set of candidates $\A_t{=}\{I^a_{t,1}, I^a_{t,2}, ... , I^a_{t,J}\}$ given by the environment. Action candidates are determined by the adjacent panos in a predefined navigation graph; each is represented by the image feature vector extracted from the perspective projection looking towards the adjacent pano. 
Selecting an action teleports the agent a few meters to the new pano. Alternatively, the agent can choose `STOP' to end the episode. On average agents have 5 actions available at each step, including `STOP'. See \cite{anderson_vision-and-language_2018} for more details.

\begin{figure*}
  \centering
  \includegraphics[trim=300 250 400 200, clip,width=0.85\linewidth]{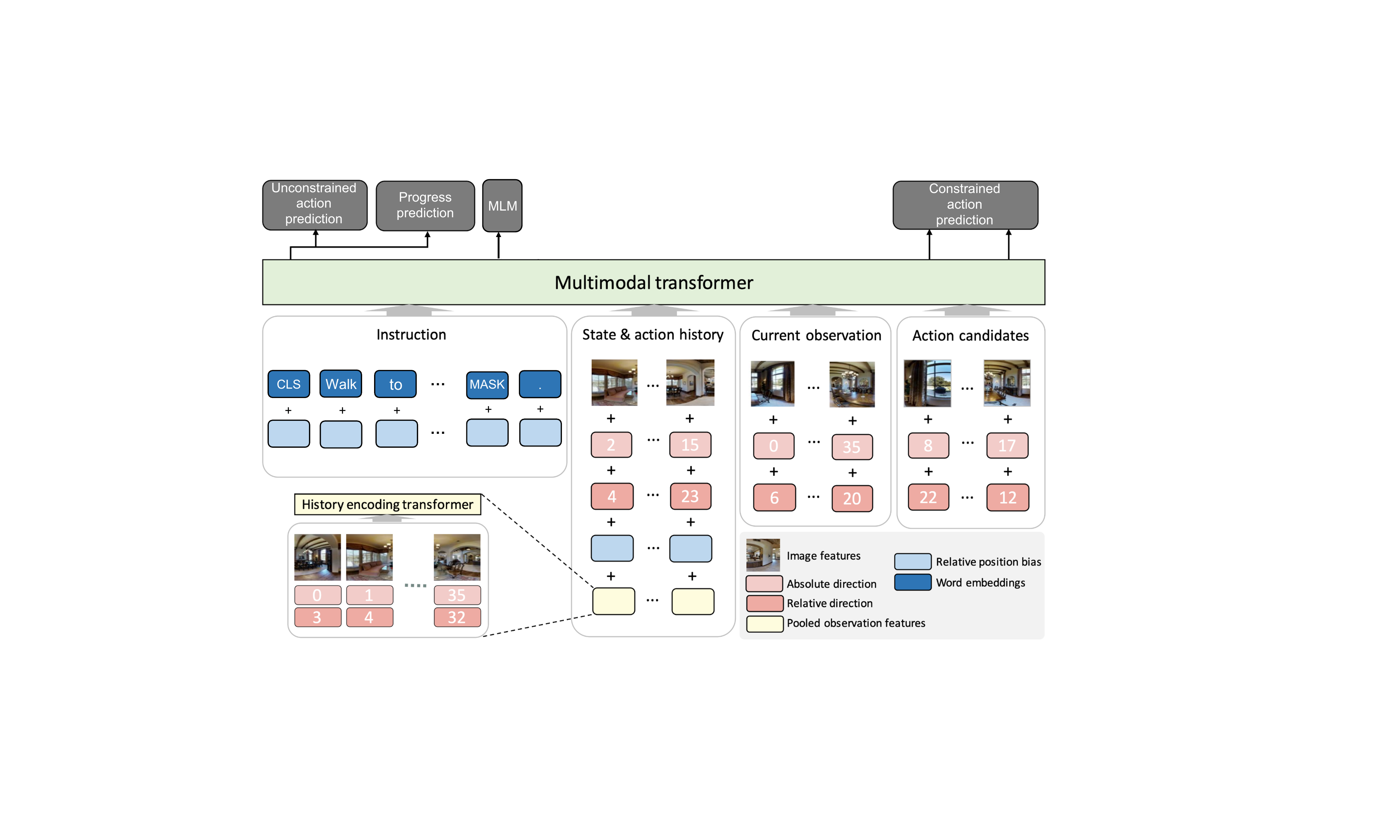}
  \caption{Agent architecture. At each time step, we combine the instruction, state and action history, current observation and action candidates into a multimodal transformer to predict the next action.
  Since observations consist of 36 image feature vectors (representing different views from a 360\degree{} camera) we compress each previous observation into a single vector, similar to \cite{chen2021hamt}.
  }
  \label{fig:model}
  \vspace{-4mm}
\end{figure*}

\noindent\textbf{Agent architecture}\quad
Our imitation-learning agent is a transformer encoder which predicts the next action $a_{t+1}$ by jointly combining all four input modalities: the instruction text $\W$, the history of observations $o_{1:t-1}$ and actions $a_{1:t-1}$, the current observation $o_t$, and the action candidates $\A_t$ (see \figref{fig:model}). At each step, all input features are concatenated into a single multimodal sequence with no attention masking, allowing every input to attend to every other input. For biasing interactions between different input modalities we include learned attention biases for each pair of input types, e.g. the instruction and the observation/action history. Like HAMT~\cite{chen2021hamt}, our approach is not autoregressive: every forward pass predicts a single action using the full history. Given our emphasis on data augmentation, we name our agent \ModelName{} for Maximum Augmentation Regime for Vision And Language navigation. Our implementation is based on mT5~\cite{xue-etal-2021-mt5}, a multilingual variant of the T5 transformer architecture~\cite{2020t5}.

\noindent\textbf{Image features}\quad
As noted above, pano observations $o_t$ and action candidates $\A_t$ are represented with sets of image features. We use precomputed, fixed 640-d features from MURAL-large \cite{mural:21}, an EfficientNet-B7 \cite{pmlr-v97-tan19a} backbone trained on 1.8B multilingual image-text pairs and 6B translation pairs. MURAL's image encoder's representational power is similar to CLIP~\cite{Radford2021LearningTV}, which is used in previous work \cite{Shen2021HowMC, chen2021hamt} and is trained on 400M English image-text pairs with a VIT~\cite{Dosovitskiy2021AnII} backbone.
To provide orientation information, each feature is combined with two learned embeddings: an \textit{absolute direction embedding} capturing the feature's orientation in the environment's fixed coordinate system, and a \textit{relative direction embedding} based on orientation relative to the agent's heading. The agent's initial heading at $t{=}0$ is given by the dataset, and is typically random. We also augment the action candidates $\A_t$ with a `STOP' action. This is convenient for modeling action classification over the candidates (refer `Action classification', below) and is represented by a zero image vector with unique direction embeddings. We use 37 absolute and relative direction embeddings, and snap features to the closest.

\noindent\textbf{Instruction encoding}\quad
The instruction $\W$ is encoded as a sequence of WordPiece~\cite{wordpiece} tokens using the mT5 vocabulary which supports up to 101 languages via a SentencePiece~\cite{kudo2018sentencepiece} model trained on mC4. Following T5, position information within the instruction is derived from relative position biases applied to the transformer's attention logits. 

\noindent\textbf{History encoding}\quad
The history of agent observations $o_{1:t-1}$ and actions $a_{1:t-1}$ is computationally expensive to process, since each pano observation $o_t$ is comprised of 36 feature vectors. Similar to \cite{chen2021hamt} we embed the 36 features from each previous pano observation into a single vector, based on the mean-pooled output of a separate transformer applied to the image features and their direction embeddings. This is added to the action candidate selected at each previous step. Position information for the state and action history is provided by relative position biases.

\begin{figure*}[t]
 \centering
 \subfloat[Agent is on the GT trajectory: Expert selects the next action in the GT trajectory.]{%
      \includegraphics[trim=0 60 0 40, clip,width=0.31\textwidth]{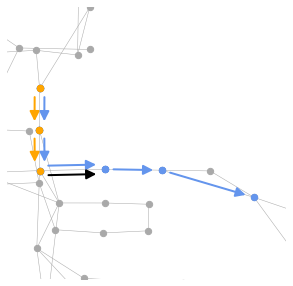}}
      \label{fig:image-a}
 \hspace{0.02\textwidth}
 \subfloat[Agent is off the GT trajectory; GT trajectory is the shortest-path from start to goal: Expert action is the first step in the recalculated shortest-path to the goal.]{%
      \includegraphics[trim=0 60 0 40, clip,width=0.31\textwidth]{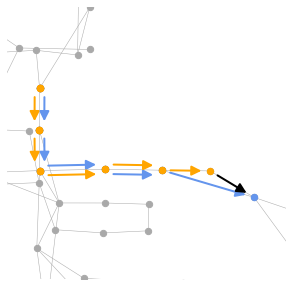}}
      \label{fig:image-b}
 \hspace{0.02\textwidth}
 \subfloat[Agent is off the GT trajectory; GT trajectory is not a shortest-path: Expert takes the shortest path back to the closest node in the GT trajectory.]{%
      \includegraphics[trim=0 50 0 50, clip,width=0.31\textwidth]{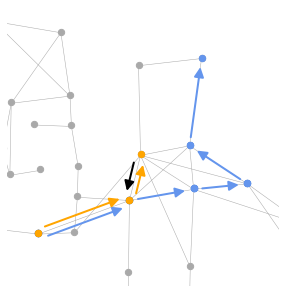}}
      \label{fig:image-c}
  \caption{Calculation of the \Dagger{} expert action (black) given the ground-truth (GT) trajectory (blue) and an agent trajectory (yellow).}
  \label{fig:expert}
  \vspace{-4mm}
\end{figure*}

\noindent\textbf{Pretraining}\quad
We train the agent in two stages. We first pretrain on a large dataset of instruction-trajectory pairs, including both model-generated instructions and trajectories containing synthesized image observations from novel viewpoints (refer \secref{sec:data}). We then finetune on a single dataset of human-annotated instruction-trajectory pairs to maximize performance on that dataset. Unlike \cite{majumdar2020improving} and \cite{Li2019RobustNW}, our transformer weights are initialized from scratch -- we do not use any image-caption datasets or text corpora to train the transformer. Since the model is not autoregressive, each training trajectory is broken down into $T$ training examples, where $T$ is the number of time steps in the trajectory. Each training example requires the model to predict the next action for a single step in a trajectory, given the full instruction $\W$, the action history $a_{1:t-1}$, the observation history $o_{1:t-1}$, the current observation $o_t$ and the set of action candidates $\A_t$. To increase the amount of supervision, during pretraining we combine four tasks:
\begin{compactitem}
\item \textbf{Masked language modeling (MLM)}~\cite{Devlin2019BERTPO}: 15\% of instruction tokens are masked, with all consecutive spans of masked tokens replaced by a single MASK token. Similar to \cite{chen2021hamt}, the model predicts the masked tokens using the surrounding text and visual clues from the observation/action history and the current observation.
\item \textbf{Progress prediction}: A small MLP is added to the output representation of the CLS token (a special symbol capturing the fused representation of the entire sequence) to predict the proportion of the trajectory that is completed (based on 20 discretized classes). Progress monitoring has been shown to improve instruction grounding \cite{ma2019selfmonitoring}.
\item \textbf{Constrained action prediction}: A classification task to predict the correct action from the constrained set of available action candidates $\A_t$. Since action candidates are inputs to the encoder (refer \figref{fig:model}), we compute the logit for each action as a learned projection of its output representation and normalize with softmax (a simplification of \cite{chen2021hamt}).
\item \textbf{Unconstrained action prediction}: A second small MLP is added to the CLS output to directly predict the next action from all 36 discretized agent-relative directions or `STOP'. Hence, these predictions are not constrained to $\A_t$, similar to the approach in \cite{hao_towards_2020}. The constrained and unconstrained action prediction tasks are highly related but complimentary; in early experiments we found that equally weighting the logits from both improves accuracy by 1-2\%, so we adopt this approach in all experiments.
\end{compactitem}

\noindent\textbf{Finetuning (behavioral cloning)}\quad
During finetuning we adapt our pretrained agent for best performance on a smaller human-annotated dataset. We update only the WordPiece embeddings in the agent and keep all other transformer weights frozen, as this makes finetuning more stable and less prone to overfitting (especially on the smaller R2R dataset). We consider two finetuning strategies. The first is \textit{behavioral cloning}. In this setting, we simply drop the instruction text masking and the MLM objective, retaining the progress prediction and constrained and unconstrained action prediction losses used in pretraining. We then finetune the agent to predict the next action at each step along ground-truth trajectories, treating imitation learning as supervised learning.  

\noindent\textbf{\Dagger training}\quad
The main weakness of behavioral cloning is that the state distribution seen in training differs from the state distribution induced by the agent during inference~\cite{ross2010efficient}. Previous works \cite{chen2021hamt, backtranslate2019} report substantial improvements by combining behavioral cloning with online reinforcement learning algorithms such as A3C~\cite{Mnih2016AsynchronousMF}. We use \textit{\Dagger}~\cite{ross2010efficient, chen2022think} to help train the agent to better recover from errors, since it is simple to implement and requires no environment interaction during training. In \Dagger{}, during each iteration of finetuning the dataset is augmented with trajectories of states visited by the current agent policy and actions given by an expert. \figref{fig:expert} explains the calculation of expert actions. We find that most of the gains are captured in a single \Dagger{} iteration.

\noindent\textbf{Pre-Exploration}\quad
While most of the focus in VLN is on instruction-following in new, unseen environments, in reality environments persist over time providing opportunities for pre-exploration. Similar to \cite{backtranslate2019,wang2018reinforced} we consider a \textit{pre-exploration} setting in which the agent may explore unseen environments with self-supervision before evaluation. Our synthetic-instruction approach is readily applicable to this scenario; we simply sample paths from the Val-Unseen or Test environments, annotate them with \Marky{} instructions, and include them in the training data.

\section{Datasets and Augmentation}
\label{sec:data}

\begin{table*}[t]
\setlength{\tabcolsep}{10pt}
\begin{center}
\small
\begin{tabularx}{\linewidth}{Xcccccc} 
 \toprule
 \textbf{Dataset} &\textbf{Instruction} & \textbf{Avg}  & \textbf{Avg} & \textbf{Environment} &\textbf{Model-}  & \textbf{Language} \\
                  &     \textbf{Count}       &\textbf{steps}        & \textbf{words} &  & \textbf{generated} & \\
    \midrule
    \midrule
Room-to-Room (R2R) \cite{anderson_vision-and-language_2018} & 14K & 5.0 & 26 & Matterport & \xmark & en \\
Room-across-Room (RxR) \cite{AlexKu:20} & 79K & 8.1 & 78 & Matterport & \xmark & en/hi/te \\
Speaker-Matterport \cite{fried2018speaker} & 178K & 5.1 & 21 & Matterport & \cmark & en \\
Marky-Matterport \cite{marky2022} & 1.0M & 9.5 & 87 & Matterport & \cmark & en/hi/te\\
Marky-Gibson (ours) & 3.2M & 7.1 & 71 & Gibson & \cmark & en/hi/te \\
 \bottomrule
\end{tabularx}
\caption{Training data. Existing datasets are situated in the 61 train environments from Matterport3D \cite{Matterport3D}. We introduce the multilingual Marky-Gibson dataset containing 3.2M model-generated navigation instructions in 491 Gibson~\cite{xiazamirhe2018gibsonenv} environments.}
\label{tab:datasets}
\vspace{-5mm}
\end{center}
\end{table*}

The datasets used for training and evaluation are described below and summarized in Table \ref{tab:datasets}.

\begin{compactitem}
\item \textbf{Room-to-Room (R2R)}~\cite{anderson_vision-and-language_2018} consists of 22K human-annotated English language navigation instructions, each describing a trajectory that traverses multiple rooms in Matterport3D \cite{Matterport3D}. This was the first dataset to use a photo-realistic environment for the instruction guided navigation task. R2R trajectories average 10m, and the trajectories are always the shortest path between the start point and the goal. 
\item \textbf{Room-across-Room (RxR)}~\cite{AlexKu:20} is a larger human-annotated dataset containing 126K instructions in English, Hindi and Telugu. To mitigate goal seeking behaviour and to ensure that agents are faithful to the instruction, RxR includes Matterport3D trajectories that are diverse in terms of length (average is 15m) and the landmarks that are referred to, and it also includes trajectories that do not go directly to the goal.
\item \textbf{Speaker-Matterport (S-MP)} \cite{fried2018speaker} is a set of 178K sampled trajectories in Matterport3D environments, annotated with synthetic instructions generated with an LSTM \cite{Hochreiter1997} Speaker model trained on R2R.  
\item \textbf{Marky-Matterport (M-MP)} \Marky{}~\cite{marky2022} is a landmark-aware multilingual instruction generator trained on RxR, used to generate 1M instructions in English, Hindi and Telugu for 330K sampled Matterport3D trajectories. 
In human wayfinding evaluations in unseen environments \Marky{} achieves close to human performance on shortest-path trajectories (e.g., R2R's paths). On the more challenging RxR paths a gap remains: human wayfinders obtain a 62\% success rate with \Marky{} vs. 78\% with human instructions.
\item \textbf{Marky-Gibson (M-Gib)} The Gibson~\cite{xiazamirhe2018gibsonenv} dataset consists of 572 indoor 3D environments. Despite its large size compared to Matterport3D, prior work has underutilized Gibson data for training VLN agents. This is primarily due to a lack of navigation trajectories and instruction annotations. To alleviate this, and unlock the Gibson dataset for VLN training, we propose an automated process to label these environments with high quality navigation graphs (see below). We then sample 1.3M trajectories and annotate them with 3.2M \Marky{} instructions in English, Hindi and Telugu. 

\end{compactitem}

\noindent\textbf{Gibson navigation graphs}\quad
In the standard VLN setting, agents are trained and evaluated using panos as observations (refer \secref{sec:approach}). Movement in the environment requires a graph with panos as nodes and edges indicating navigability. Navigation graphs for Matterport3D were generated by \cite{anderson_vision-and-language_2018}, using a semi-automated process combined with human visual inspection. However, there are no navigation graphs for Gibson environments and the size of the dataset precludes human inspection. We therefore train a model on panos and navigation graphs from the Matterport3D train split to classify whether a patch of pixels in a pano constitutes a navigable direction. The model is based on RedNet \cite{jiang2018rednet}, an RGB-D encoder-decoder first proposed for image segmentation, using a ResNet-50 \cite{he2015deep} backbone. The output space is discretized into $8 \times 16 \times 5$ pitch, heading and distance buckets. During training each bucket is assigned a positive value if the corresponding location corresponds to a navigable node, and 0 otherwise.

To compute Gibson navigation graphs, we combine model edge predictions with obstacle information from the dataset's 3D meshes. We add an edge between pano nodes $i$ and $j$ if the following boolean expression evaluates to true:
\begin{equation*}
\label{eq:graph}
   e(i, j)=(\lambda_d \frac{g_{i,j}}{s_{i,j}} - \lambda_p p_{i,j} \leq 1)  \land (s_{i,j} \leq 3.5)   \land  (|z_i - z_j| \leq 3) 
\end{equation*}
\noindent
where $g_{i,j}$ is the geodesic distance (accounting for obstacles) between nodes $i$ and $j$ calculated using the Habitat Simulator \cite{habitat19iccv}, $s_{i,j}$ is the straight-line Euclidean distance between nodes $i$ and $j$, $p_{i,j}$ is the model probability of an edge connecting nodes $i$ and $j$, $z_i$ is the vertical coordinate of pano $i$, and $\lambda_d$ and $\lambda_p$ are weighting parameters. The first term captures model predictions and encourages edges between panos that have few intervening obstacles. The second term ensures that nodes are within 3.5m, and the third term ensures that nodes are within 3m in the vertical axis (these values are chosen based on \cite{anderson_vision-and-language_2018}). Finally, to ensure that the navigation graph for each environment is fully connected, we compute the minimum spanning tree (MST)~\cite{kruskal1956shortest} of the graph with the edge weights given by the first term in the equation for $e(i, j)$, and apply a logical `OR' operation over $e(i,j)$ and the MST.

To set the weighting parameters $\lambda_d$ and $\lambda_p$, we perform grid search to maximize the $F_1$ score when predicting edges in Matterport3D val environments. $F_1$ uses the standard
calculation where the population includes all pairs of nodes
in each environment, and the true condition is positive if an
edge exists in the hand-crafted navigation graphs from \cite{anderson_vision-and-language_2018}.
Our approach achieves 
an $F_1$ score of 0.70, precision of 0.695, and recall of 0.713. The average edge length in the generated Gibson graphs is 3.02m (median of 2.06m), and the average node degree is 4.15 (median of 4).

\noindent\textbf{Trajectory sampling and instruction generation}\quad
Using the generated navigation graphs, we sample trajectories from 491 Gibson train and val environments (we do not use test environments). 
Unlike Matterport3D, Gibson lacks room annotations, which precludes us from using the two-step sampling approach from RxR. Instead, we use a simpler approach: we randomly sample 3 panos, and use a TSP solver to find the shortest path that visits all 3 panos. Trajectories longer than 40m or 16 steps are discarded, and no more than 3K paths are sampled per environment. This procedure generates 1.06M paths, with an average of 7.1 steps and length of 19.3m. Using \Marky{} we annotate each trajectory with English, Hindi and Telugu instructions to create the Marky-Gibson dataset.

\begin{table*}[t]
\setlength{\tabcolsep}{5pt}
\begin{center}
\small
\begin{tabularx}{\linewidth}{lXccccccccccccccc}
& & \multicolumn{6}{c}{\textbf{Pretraining Data}}    &   & \multicolumn{4}{c}{\textbf{RxR \textsc{Val-Unseen}}} &  & \multicolumn{3}{c}{\textbf{R2R \textsc{Val-Unseen}}}         \\ \cmidrule{3-8} \cmidrule{10-13} \cmidrule{15-17} 
& \textbf{Size} & \textbf{R2R} & \textbf{RxR} & \textbf{S-MP} & \textbf{M-MP} & \textbf{M-Gib} & \textbf{SE3DS} & \textbf{Iterations} & \textbf{NE} & \textbf{SR} & \textbf{NDTW} & \textbf{SDTW} & \textbf{} & \textbf{NE} & \textbf{SR}  & \textbf{SPL} \\
\midrule
1 & Base & \cmark & \cmark & & & & & 630K & 11.16 & 23.2 & 40.1 & 19.0 & & 7.58 & 33.5 & 31.9\\
2 & Base & \cmark & \cmark & \cmark & & & & 1.68M & 11.17 & 25.0 & 40.3 & 21.0 & & 6.50 & 41.0 & 39.0\\
3 & Base & \cmark & \cmark & \cmark & \cmark & & & 2.94M & 7.53 & 45.0 & 56.9 & 39.2 & & 7.07 & 40.3 & 37.2 \\
4 & Base & \cmark & \cmark & & \cmark & & & 2.00M & 6.90 & 50.2 & 60.3 & 43.9 & & 6.07 & 50.0 & 46.5\\
5 & Base & \cmark & \cmark & \cmark & \cmark & & \cmark & 1.94M & 6.68	& 50.9	& 61.6 & 45.0 & & 5.34 & 52.2 & 49.0\\
6 & Base & \cmark & \cmark & \cmark & \cmark & \cmark & & 3.00M & 5.90 & 55.6 & 64.4 & 49.0 & & 5.24 & 52.5 & 47.9\\
7 & Base & \cmark & \cmark & \cmark & \cmark & \cmark & \cmark & 2.80M & 5.75 & 55.9 &	65.1 & 49.3 & & 5.08 & 54.2 & 50.1\\
8 & Large & \cmark & \cmark & \cmark & \cmark & \cmark & \cmark & 4.80M & 5.72 & 58.3 & 66.1 & 51.7 & & 4.87 & 55.6 & 52.8 \\
9 & Large & \cmark & \cmark &  & \cmark & \cmark & \cmark & 5.14M & \textbf{5.56} & \textbf{59.4} & \textbf{67.0} & \textbf{52.7} & & \textbf{4.84} & \textbf{57.4} & \textbf{54.6}\\
\bottomrule
\end{tabularx}
\caption{Comparison of pretraining settings. Best results (without finetuning) are obtained by combining the \textbf{R2R} and \textbf{RxR} datasets \textit{with} Marky-generated instructions in both Matterport3D (\textbf{M-MP}) and Gibson (\textbf{M-Gib}) environments, \textit{without} Speaker-generated instructions (\textbf{S-MP}), and \textit{with} synthesis of observations from novel viewpoints using \textbf{SE3DS} (row 9).}
\label{tab:pretraining}
\vspace{-7mm}
\end{center}
\end{table*}

\noindent\textbf{Synthesizing image observations with SE3DS}\quad
One weakness of training VLN agents on pano images is that training trajectories are constrained to the locations of the captured images. VLN agents tend to overfit to these trajectories \cite{Zhang2020DiagnosingTE}, contributing to a performance drop in unseen environments. \cite{koh2021pathdreamer,koh2022simple} showed that a strong generative model is capable of successfully rendering high resolution panos from novel viewpoints, and that training VLN agents with spatially-perturbed panos could improve the success rate of the agent on R2R Val-Unseen by 1.5\%. To assess if this approach is complimentary to instruction augmentation, we use the proposed SE3DS (Simple and Effective 3D Synthesis) model  
to augment panoramas from the Matterport environments. Following \cite{koh2022simple}, we create 200 variations of each environment which are randomly sampled during training. In each environment variation, with 50\% probability a pano will be spatially-perturbed by up to 1.5m and re-rendered at the new location using SE3DS.

\section{Experiments}
\label{sec:experiments}

\noindent\textbf{Pretraining settings}\quad
In \tabref{tab:pretraining} we explore pretraining using varying amounts and types of augmented data. During pretraining, we monitor one-step action prediction accuracy on ground-truth trajectories using held-out instructions from RxR and R2R Val-Unseen. Each setting is trained until convergence, requiring more iterations (\textbf{Its}) for larger models and datasets. We select the best checkpoint based on one-step prediction accuracy then perform a full evaluation using standard VLN path-fidelity metrics \cite{anderson_evaluation_2018,magalhaes2019effective}: Navigation Error (\textbf{NE} $\downarrow$, the average distance in meters between the agent's final position and the goal), Success Rate (\textbf{SR}$\uparrow$, the proportion of trajectories with NE $<$ 3m), Success rate weighted by normalized inverse Path Length (\textbf{SPL} $\uparrow$), normalized dynamic time warping (\textbf{NDTW}$\uparrow$), and success weighted DTW (\textbf{SDTW}$\uparrow$).

\noindent\textbf{Speaker vs. \Marky{}}\quad
Consistent with previous work, we find that data augmentation with synthetic instructions from the \cite{fried2018speaker} Speaker model improves performance (row 2 vs. 1, +2\% SR on RxR and +8\% on R2R), but instructions from \Marky{} \cite{marky2022} are far more effective (row 4 vs. 1, +27\% SR on RxR and +17\% on R2R). This is consistent with human evaluations of instruction quality, confirming that improvements in instruction-generation flow through to instruction-following performance. Interestingly, we find that combining the Speaker model with Marky leads to worse performance on both RxR and R2R (row 3 vs. 4, and also row 8 vs. 9), which we attribute to the introduction of noise from the lower-quality Speaker instructions.

\noindent\textbf{Gibson, SE3DS and model size}\quad
Augmentation with \Marky{} instructions in Gibson environments (row 6 v. 3) provides a substantial boost (+11\% SR on RxR and +12\% on R2R), suggesting that the returns from scaling synthetic instructions to more environments are not exhausted. Using SE3DS to synthesize image observations from novel viewpoints improves +6\% SR on RxR and +12\% on R2R (row 5 vs. 3), but this benefit is substantially reduced (+0\% SR on RxR and +2\% on R2R, row 7 vs. 6) if Gibson is included, presumably because new environments also increase viewpoint variety. Most experiments use the mT5-base \cite{xue-etal-2021-mt5} model; switching to mT5-large provides a further performance boost (+2\% SR on RxR and +1\% on R2R, row 8 vs. 7). Our best pretraining results on both RxR and R2R are achieved using an mT5-large model with all the previously mentioned data, but leaving out the Speaker instructions (row 9). We use this checkpoint in all finetuning experiments. This agent pretrains for 5.14M iterations, which, using a batch size of 128, represents over 650M steps of experience (over 700M including finetuning). 

\begin{table*}[t]
\setlength{\tabcolsep}{4.0pt}
\begin{center}
\small
\begin{tabularx}{\linewidth}{Xcccccccccccccc} 
 &  \multicolumn{4}{c}{\textbf{\textsc{Val-Seen}}} & & \multicolumn{4}{c}{\textbf{\textsc{Val-Unseen}}} & & \multicolumn{4}{c}{\textbf{\textsc{Test (Unseen)}}} \\
\cmidrule{2-5} \cmidrule{7-10} \cmidrule{12-15}
\textbf{Agent} & \textbf{NE} & \textbf{SR} & \textbf{NDTW} & \textbf{SDTW} & & \textbf{NE} & \textbf{SR} & \textbf{NDTW} & \textbf{SDTW} & & \textbf{NE} & \textbf{SR} & \textbf{NDTW} & \textbf{SDTW} \\
\midrule
LSTM \cite{AlexKu:20} & 10.7 & 25.2 & 42.2 & 20.7  & & 10.9 & 22.8 & 38.9 & 18.2 && 12.0 & 21.0 & 36.8 & 16.9\\
EnvDrop+ \cite{Shen2021HowMC} & - & - & - & -  & & - & 42.6 & 55.7 & - && - & 38.3 & 51.1 & 32.4\\
CLEAR-C \cite{li2022clear} & - &- &- & - && - & - & - & - && - & 40.3 & 53.7 & 34.9 \\
HAMT \cite{chen2021hamt}  & - & 59.4 & 65.3 & 50.9  && - & 56.5  & 63.1 & 48.3 && 6.2 & 53.1 & 59.9 & 45.2\\
EnvEdit* \cite{Li2022EnvEditEE} & - & 67.2 & 71.1 & 58.5 && - & 62.8 & 68.5 & 54.6 && \textbf{5.1} & 60.4 & 64.6 & 51.8 \\
\ModelName (Pretrained) & 3.62 & 72.7 & 77.0 & 65.9 & & 5.56 & 59.4 & 67.0 & 52.7 & & - & - & - & -\\
\ModelName (Finetuned-BC) & 3.25 & 75.4 & 79.0 & 68.7 & & 4.80 & 63.7 & 70.6 & 56.9 & & - & - & - & -\\
\ModelName (\Dagger) & \textbf{3.01} & \textbf{75.9} & \textbf{79.1} & \textbf{68.8} & & \textbf{4.49} & \textbf{64.8} & \textbf{70.8} & \textbf{57.5} & & 5.5 & \textbf{60.7} & \textbf{66.8} & \textbf{53.5}\\
\midrule
\ModelName (Pre-Explore)$\dagger$ & 3.33 & 73.7 & 77.6 & 66.6 && 4.19 & 66.5 &  72.2 & 59.1 && 5.2 & 61.8 & 68.6 & 54.8 \\
\midrule
Human \cite{AlexKu:20} & - & - & - & - && - & - &  - & - && 0.9 & 93.9 & 79.5 & 76.9 \\
\bottomrule
\multicolumn{12}{l}{\scriptsize{*Results from an ensemble of three agents.}}
\end{tabularx}
\caption{Results on RxR. Our \ModelName{} agent trained with imitation learning -- behavioral cloning (BC) or \Dagger{} -- outperforms all existing RL agents.   Pre-Exploration in the eval environments ($\dagger$ a form of privileged access, but still without human annotations) can provide a further boost.}
\label{tab:RxR}
\vspace{-5.5mm}
\end{center}
\end{table*}

\begin{table*}[t]
\setlength{\tabcolsep}{5.0pt}
\begin{center}
\small
\begin{tabularx}{\linewidth}{Xcccccccccccccc} 
 &  \multicolumn{4}{c}{\textbf{\textsc{Val-Seen}}} & & \multicolumn{4}{c}{\textbf{\textsc{Val-Unseen}}} & & \multicolumn{4}{c}{\textbf{\textsc{Test (Unseen)}}}\\
 \cmidrule{2-5} \cmidrule{7-10} \cmidrule{12-15}
\textbf{Agent}    & \textbf{TL}  & \textbf{NE} &  \textbf{SR} & \textbf{SPL} && \textbf{TL}  & \textbf{NE} &  \textbf{SR} & \textbf{SPL} && \textbf{TL}  & \textbf{NE} &  \textbf{SR} & \textbf{SPL}   \\
 \midrule
PREVALENT \cite{hao_towards_2020} & 10.32 & 3.67 & 69\pz & 65\pz && 10.19 & 4.71 & 58\pz & 53\pz && 10.51 & 5.30 & 54 & 51 \\
RecBERT \cite{Hong2021VLNBERTAR} & 11.13 & 2.90  & 72\pz & 68\pz  && 12.01  & 3.93 &  63\pz &  57\pz  && 12.35 &  4.09 &  63  & 57 \\
EnvDrop+ \cite{Shen2021HowMC} & - & - & - & - && - & - &  59.2 & 52.9 && - & - & 59 & 53 \\
AirBERT \cite{Guhur2021AirbertIP} & 11.09 & 2.68 & 75\pz & 70\pz && 11.78 & 4.01 & 62\pz & 56\pz && 12.41 & 4.13 & 62 & 57 \\
HAMT \cite{chen2021hamt} & 11.15  & 2.51  & 76\pz & 72\pz  && 11.46 &  2.29 &  66\pz &  61\pz &&  12.27 &  3.93  & 65  & 60 \\
REM \cite{Liu2021VisionLanguageNW} & 10.88 & 2.48 & 75.4 & 71.8 && 12.44 & 3.89 & 63.6 & 57.9 && 13.11 & 3.87 & 65 & 59 \\
EnvEdit* \cite{Li2022EnvEditEE} & 11.18 & \textbf{2.32} & \textbf{76.9} & \textbf{73.9} && 11.13 & \textbf{3.24} & \textbf{68.9} & \textbf{64.4} && 11.90 & \textbf{3.59} & \textbf{68} & \textbf{64} \\
\ModelName (Pretrained) & 10.32 & 3.73 & 68.2 & 64.9 && 9.71 & 4.84 & 57.4 & 54.6 && - & - & - & - \\
\ModelName (Finetuned-BC) & 10.43 & 3.11 & 72.3 & 68.9 && 9.72 & 4.20 & 63.0 & 60.0 && - & - & - & - \\
\ModelName (\Dagger) & 10.60 & 2.99 & 73.0 & 69.1 && 10.15 & 4.06 & 64.8 & 60.7 && 10.22 & 4.18 & 62 & 58 \\
\midrule
Human \cite{anderson_vision-and-language_2018} & - & - & - & - && - & - &  - & - && 11.90 & 1.61 & 86 & 76\pz \\
\bottomrule
\multicolumn{12}{l}{\scriptsize{*Results from an ensemble of three agents.}}
\end{tabularx}
\caption{Results on R2R. \ModelName{} achieves strong performance but not state-of-the-art, which we attribute to domain differences between the R2R and RxR (which was used to train \Marky{}).}
\label{tab:R2R}
\vspace{-5mm}
\end{center}
\end{table*}

\noindent\textbf{Finetuning}\quad
In Tables \ref{tab:RxR} and \ref{tab:R2R} we compare results for our \ModelName{} agent after finetuning to previous work on the RxR and R2R datasets. On both datasets, finetuning with behavioral cloning on just human-annotated data (Finetuned-BC) substantially improves the pretrained model. The improvement from using \Dagger{} over behavioral cloning is small but consistent. On the RxR dataset, \ModelName{} outperforms all prior work. Evaluating on novel instruction-trajectories in seen environments (Val-Seen), we improve over the state-of-the-art by 8\%, reaching 79.1 NDTW. In new, unseen environments (Test), we improve by 2\%, achieving 66.8 NDTW. Self-training with \Marky{} synthetic instructions in the Test environments (a form of privileged access, but still without human annotations) improves performance by an additional 2\% to 68.6 NDTW.

\noindent\textbf{RxR vs. R2R}\quad
On the English-only R2R dataset (\tabref{tab:R2R}), \ModelName{} achieves strong performance but not state-of-the-art. Surprisingly, the Val-Unseen success rate (SR) of 64.8\% is the same for both RxR and R2R, whereas typically RxR performance is lower since the trajectories are longer and more varied. Noting that \Marky{} was trained on RxR, we attribute lower relative performance on R2R to domain differences between R2R and RxR. While the average length of instructions in R2R is 26 words, RxR has an average of 87 words --- 3 times more. This is partly because RxR instructions are more verbose, often describing objects in more detail and including state verification. Further, cultural differences arising from the data collection process (annotators from USA or from India) may also contribute to the distribution shift due to subtle differences in the vocabulary and structure of language used to form the instructions. 
We note however, that while our augmentation approach focuses on scaling up in terms of high quality instructions, EnvEdit \cite{Li2022EnvEditEE} focuses on generalization through augmentation of visual features. These two approaches may ultimately prove to be complementary.

\section{Conclusion}

We build a purely imitation learning agent that achieves state-of-the-art results on the RxR benchmark. This result paves a new path towards improving instruction-following agents, emphasizing large-scale imitation learning with generic architectures, along with a focus on developing synthetic instruction generation capabilities -- which are shown to directly improve instruction-following performance. We find that aligning synthetic instructions to the target domain is essential, as seen through the gap in performance on R2R. On RxR, the performance improvement over the state-of-the-art is much larger in seen environments (+8\%) than unseen test environments (+2\%). Scaling to even more indoor environments might improve generalization further.

{\small
\bibliographystyle{ieee_fullname}
\bibliography{main}
}

\clearpage

\begin{figure*}[h]
  \centering
  \includegraphics[width=0.63\linewidth]{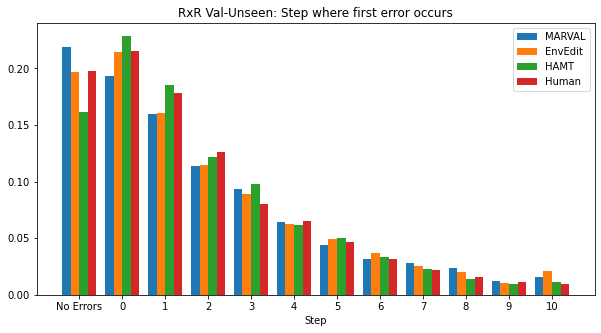}
  \vspace{-2mm}
  \caption{Error analysis indicating at which trajectory step each agent makes its first mistake. Surprisingly, \ModelName{} makes less errors (produces more perfect trajectories) than the prior work \textit{and human followers}. Since human followers still significantly outperform \ModelName{} overall, this suggests the main focus for future agent improvement should be on recovering from errors.}
  \label{fig:errors}
\end{figure*}

\appendix

\section{Appendix}

\subsection{Limitations and Future Work}

As we discuss in the main paper, our approach achieves strong but not state-of-the-art results on the R2R dataset, which
we attribute to domain differences between R2R and RxR (noting that the \Marky{} instruction generator we use for data augmentation was trained on RxR). One way to address this limitation would be by re-training \Marky{} on R2R data, although this would face some hurdles since R2R lacks the annotator pose traces that were used by \Marky{} when training on RxR. 

To better understand the failure modes of our approach, in \figref{fig:errors} we plot a distribution indicating the step in the trajectory where an agent makes its first error. We analyze the RxR Val-Unseen split and compare \ModelName{} to the previous state-of-the-art approaches, EnvEdit and HAMT, as well as human instruction-following demonstrations from the RxR dataset. \ModelName{} makes fewer errors than the prior approaches, especially at the start of the trajectory, but also \textit{fewer errors than human followers}. Since human followers still significantly outperform \ModelName{} overall -- in terms of navigation error (0.79m vs. 4.49m), success rate (94.5 vs. 64.8) and path-fidelity metrics such as NDTW (81.8 vs. 70.8) -- this suggests the main focus for future agent improvement should be on recovering from errors, which human wayfinders clearly do extremely well in order to still reach the goal in 94.5\% of episodes.

\subsection{Implementation Details}

\textbf{Pretraining}\quad
In all experiments we train with a batch size of 128 using the AdaFactor optimizer. During pretraining, we use dropout of 0.1 and a learning rate that exponentially-decays from 0.1. We monitor one-step action prediction accuracy on ground-truth trajectories using held-out instructions from unseen environments (RxR Val-Unseen and R2R Val-Unseen). We pretrain until convergence and then select the best snapshot based on one-step action prediction accuracy on RxR Val-Unseen.

\textbf{Finetuning}\quad
During finetuning, we use a constant learning rate of 0.001 and dropout of 0.2. Since the human-annotated datasets used for finetuning (RxR and R2R) are relatively small, during finetuning we update only the WordPiece embeddings in the agent and keep all other transformer weights frozen. This makes finetuning more stable and less prone to overfitting (especially on the smaller R2R dataset). We finetune for a maximum of 150K iterations while monitoring standard VLN path-fidelity metrics such as success rate (SR). We select the best snapshot based on SR on Val-Unseen.

\subsection{Additional Pretraining Results}

In \tabref{tab:pretraining-val-seen} we report pretraining results on the \textit{Val-Seen} splits, complementing the \textit{Val-Unseen} results included in the main paper. We observe the same trends in the seen environments as in the new, unseen environments (Val-Unseen), although the relative improvement from using a larger model (row 8 vs. 7) is larger.

\begin{table*}[h]
\setlength{\tabcolsep}{4.5pt}
\begin{center}
\small
\begin{tabularx}{\linewidth}{lXccccccccccccccc}
& & \multicolumn{6}{c}{\textbf{Pretraining Data}}    &   & \multicolumn{4}{c}{\textbf{RxR \textsc{Val-Seen}}} &  & \multicolumn{3}{c}{\textbf{R2R \textsc{Val-Seen}}}         \\ \cmidrule{3-8} \cmidrule{10-13} \cmidrule{15-17} 
& \textbf{Size} & \textbf{R2R} & \textbf{RxR} & \textbf{S-MP} & \textbf{M-MP} & \textbf{M-Gib} & \textbf{SE3DS} & \textbf{Iterations} & \textbf{NE} & \textbf{SR} & \textbf{NDTW} & \textbf{SDTW} & \textbf{} & \textbf{NE} & \textbf{SR} & \textbf{SPL} \\
\midrule
1 & Base & \cmark & \cmark & & & & & 630K & 12.14 & 22.3 & 39.3 & 18.8 & & 7.45 & 37.9 & 36.3\\
2 & Base & \cmark & \cmark & \cmark & & & & 1.68M & 11.52 & 26.7 & 42.4 & 22.9 & & 5.19 & 52.6 & 50.5\\
3 & Base & \cmark & \cmark & \cmark & \cmark & & & 2.94M & 7.23 & 48.5 & 59.3 & 42.4 & & 6.19 & 48.7 & 45.7 \\
4 & Base & \cmark & \cmark & & \cmark & & & 2.00M & 4.42 & 64.8 & 73.2 & 58.5 & & 3.86 & 67.2 & 64.2\\
5 & Base & \cmark & \cmark & \cmark & \cmark & & \cmark & 1.94M & 5.15	& 62.0	& 69.9 & 55.6 & & 4.61 & 59.2 & 56.2\\
6 & Base & \cmark & \cmark & \cmark & \cmark & \cmark & & 3.00M & 5.09 & 61.4 & 68.7 & 51.2 & & 4.77 & 57.5 & 53.0\\
7 & Base & \cmark & \cmark & \cmark & \cmark & \cmark & \cmark & 2.80M & 5.22 & 60.0 &	68.3 & 53.1 & & 4.70 & 59.6 & 55.1\\
8 & Large & \cmark & \cmark & \cmark & \cmark & \cmark & \cmark & 4.80M & 3.85 & 71.0 & 75.9 & 64.3 & & 4.05 & 66.1 & 62.6 \\
9 & Large & \cmark & \cmark &  & \cmark & \cmark & \cmark & 5.14M & \textbf{3.62} & \textbf{72.7} & \textbf{77.0} & \textbf{65.9} & & \textbf{3.73} & \textbf{68.2} & \textbf{64.9}\\
\bottomrule
\end{tabularx}
\caption{Comparison of pretraining approaches reporting Val-Seen results on RxR and R2R. }
\label{tab:pretraining-val-seen}
\end{center}
\end{table*}

\subsection{Performance by Language}

In \tabref{tab:RxR_lang} we report results by language on the RxR Val-Seen and Val-Unseen splits in comparison to the previous state-of-the-art EnvEdit model. Improvements are across-the-board and results on each language are similar.

\begin{table*}[h]
\setlength{\tabcolsep}{4.0pt}
\begin{center}
\begin{tabularx}{\linewidth}{Xcccccccccc} 
 &  \multicolumn{4}{c}{\textbf{\textsc{Val-Seen}}} & & \multicolumn{4}{c}{\textbf{\textsc{Val-Unseen}}}  \\
\cmidrule{2-5} \cmidrule{7-10} 
\textbf{Agent} & \textbf{NE} $\downarrow$ & \textbf{SR}$\uparrow$ & \textbf{NDTW}$\uparrow$ & \textbf{SDTW}$\uparrow$ & & \textbf{NE}$\downarrow$ & \textbf{SR}$\uparrow$ & \textbf{NDTW}$\uparrow$ & \textbf{SDTW}$\uparrow$ \\
\midrule
\textbf{English (en-IN):} & \multicolumn{8}{l}{} \\
EnvEdit* \cite{Li2022EnvEditEE} & 3.82 & 68.01 & 71.45 & 59.09 && 4.42 & 62.03  & 67.89 &  54.15 \\
\ModelName & 3.10  & 75.77 & 78.79 & 68.86  &&  4.49 & 64.83 & 70.67 &  57.64 \\
\midrule
\textbf{English (en-US):} & \multicolumn{8}{l}{} \\
EnvEdit* \cite{Li2022EnvEditEE} & 4.22 & 66.19  &  69.52 & 56.95 && 4.33 & 61.70 &  67.56 &  52.94 \\
\ModelName & 3.53 & 72.22  & 76.19 & 64.62  && 4.46 & 64.47  & 70.28 &  56.46  \\
\midrule
\textbf{Telugu (te-IN):} & \multicolumn{8}{l}{} \\
EnvEdit* \cite{Li2022EnvEditEE} & 3.72 & 65.59  & 70.70 & 57.37  && 4.49 & 61.85 & 67.84 & 53.75  \\
\ModelName & 2.82 & 76.38 & 79.34 & 69.43 && 4.56 & 63.85  & 69.92  & 56.30   \\
\midrule
\textbf{Hindi (hi-IN):} & \multicolumn{8}{c}{} \\
EnvEdit* \cite{Li2022EnvEditEE} & 4.01 & 68.56  & 71.90 & 59.92 && 4.20 & 64.54  & 69.74 &   56.41\\
\ModelName &2.95  &76.61& 80.05 & 69.52 && 4.30 & 65.85  & 72.02 & 58.88  \\
\bottomrule
\multicolumn{8}{l}{\scriptsize{*Results from an ensemble of three agents.}}
\end{tabularx}
\caption{Breakdown of the results on RxR for each language for our best model and the best performing previous model from \cite{Li2022EnvEditEE}. Predicted paths for EnvEdit were provided by the authors.}
\label{tab:RxR_lang}
\end{center}
\end{table*}

\subsection{Marky Synthetic Instruction Examples}
\begin{figure*}[h]
  \centering
  \textbf{Marky Instruction:} You are facing a potted plant. Turn around, exit the room. Turn right, walk straight. Turn right, you can see stairs. Get down through the stairs. Turn right, get down through the stairs. Stand on the second step of the stair from the top. This is the end point.
  \vspace{1mm}\\
  \includegraphics[width=0.48\linewidth]{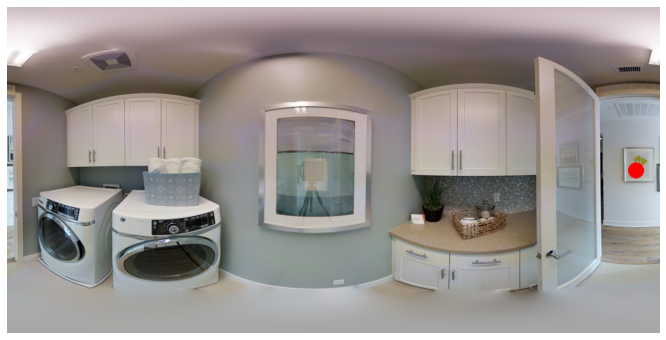}
  \includegraphics[width=0.48\linewidth]{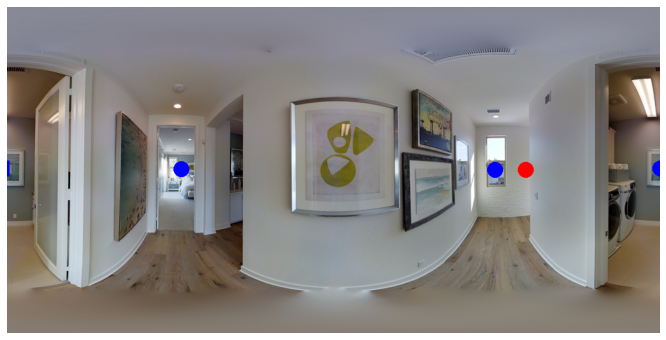} \vspace{2mm}\\
  \includegraphics[width=0.48\linewidth]{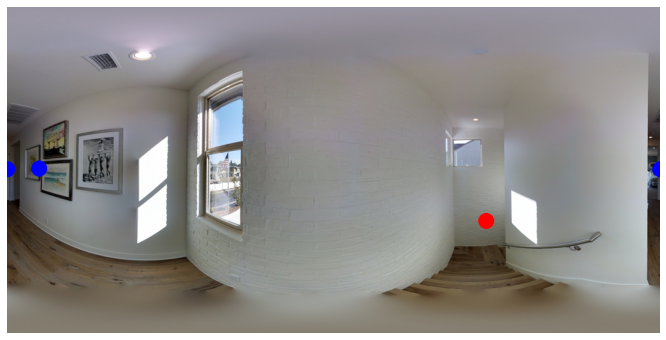}
  \includegraphics[width=0.48\linewidth]{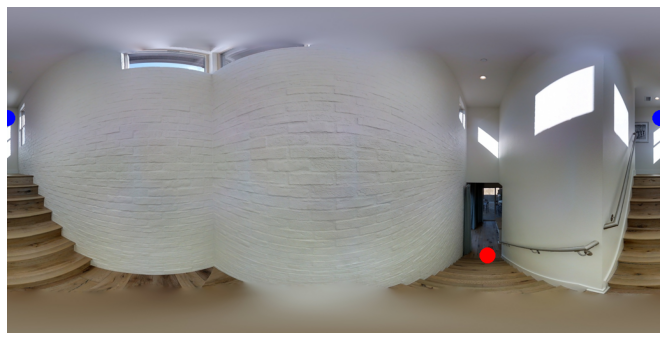} \vspace{2mm}\\
  \includegraphics[width=0.48\linewidth]{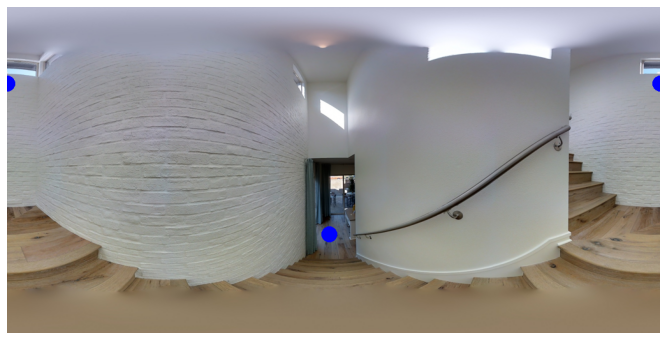}
  \caption{Example Marky (synthetic) instruction for a sampled trajectory. The images are 360\degree{} panoramas rotated so that the direction faced by the agent is the in center. \textcolor{blue}{Blue} dots indicate directions the agent can move in the underlying navigation graph. The correct action at each step is colored in \textcolor{red}{red}. Note that the potted plant mentioned in the instruction is on the countertop. }
  \label{fig:marky1}
\end{figure*}

\begin{figure*}[h]
  \centering
  \textbf{Marky Instruction:} You begin facing a living space, take a step forward and then behind the brown chair that's in front of you, there is a archway, go through that archway, and then go down the hallway to the left, turn into the first room on the right, it is a bathroom, take a step inside and you're done.
  \vspace{1mm}\\
  \includegraphics[width=0.48\linewidth]{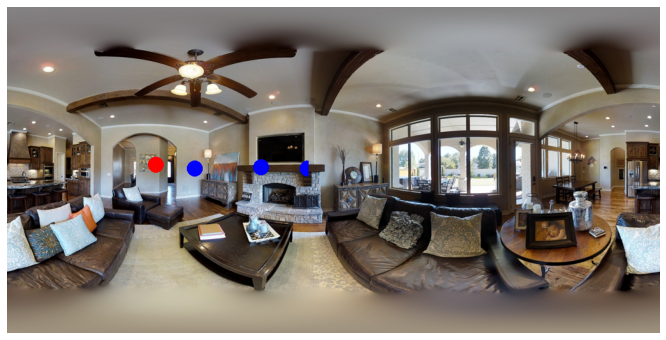}
  \includegraphics[width=0.48\linewidth]{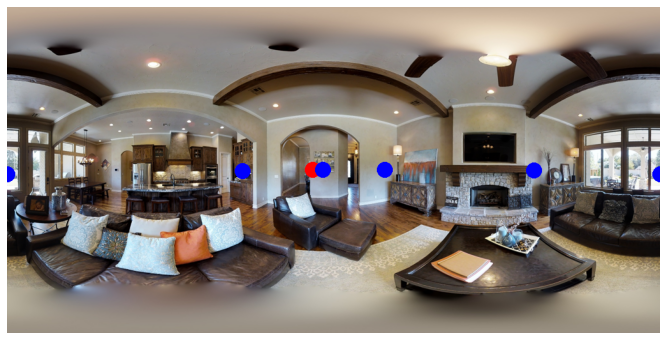} \vspace{2mm}\\
  \includegraphics[width=0.48\linewidth]{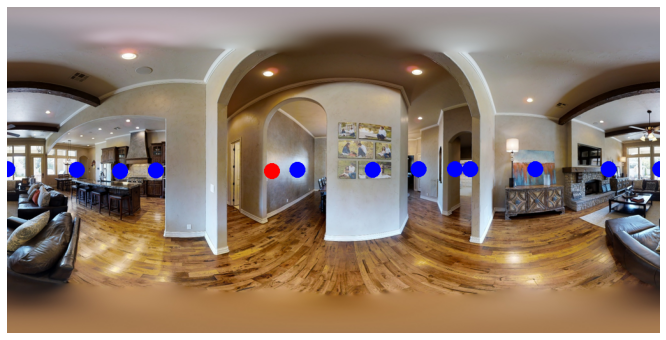}
  \includegraphics[width=0.48\linewidth]{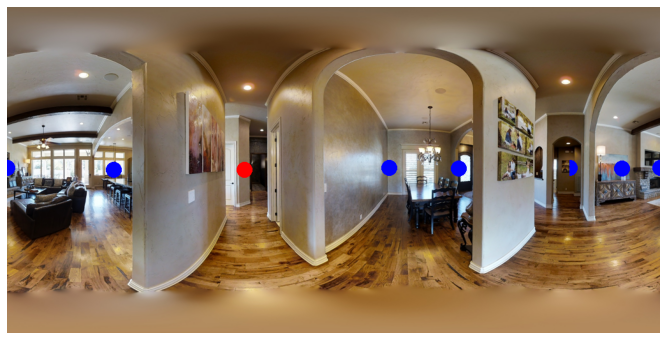} \vspace{2mm}\\
  \includegraphics[width=0.48\linewidth]{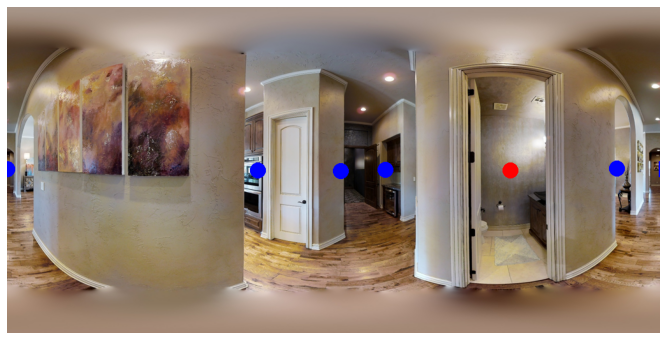}
  \includegraphics[width=0.48\linewidth]{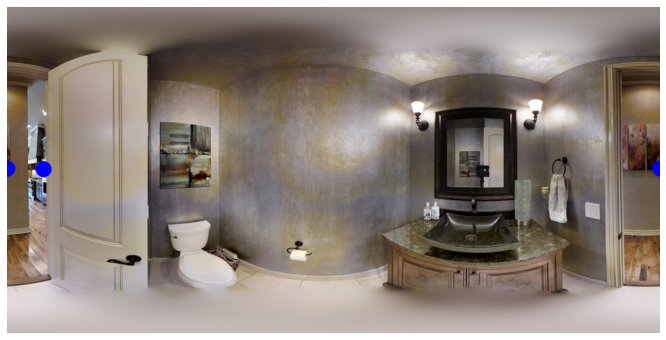}
  \caption{Example Marky (synthetic) instruction for a sampled trajectory. The images are 360\degree{} panoramas rotated so that the direction faced by the agent is the in center. \textcolor{blue}{Blue} dots indicate directions the agent can move in the underlying navigation graph. The correct action at each step is colored in \textcolor{red}{red}.}
  \label{fig:marky2}
\end{figure*}

Figures \ref{fig:marky1} and \ref{fig:marky2} provide examples of Marky-generated (synthetic) instructions instruction and their associated trajectories.

\subsection{MARVAL Instruction-Following Examples}

\begin{figure*}[h]
  \centering
  \textbf{Instruction:} You're facing towards a wooden shelf, turn slight left and move forward you can see a gas stove in front of you walk near the gas stove you can see a open door in front of you, turn right and exit the room by moving one step forward, turn slight left and you can see a stair case in front of you walk near the stair case, turn slight right and you can see an open door in front of you walk near the open door, turn slight right and you can see a washing machine in front of you walk near the washing machine and stand near the washing machine and this will be your end point.   \vspace{1mm}\\
  \includegraphics[width=0.48\linewidth]{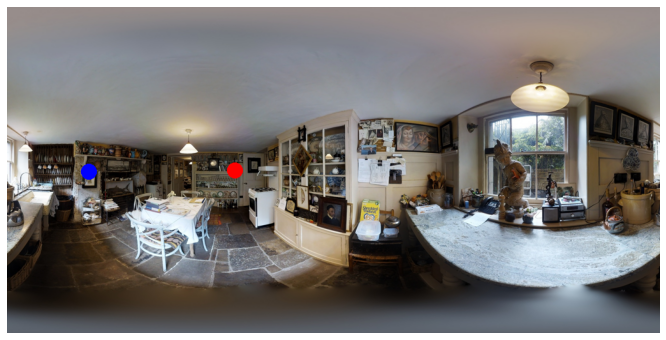}
  \includegraphics[width=0.48\linewidth]{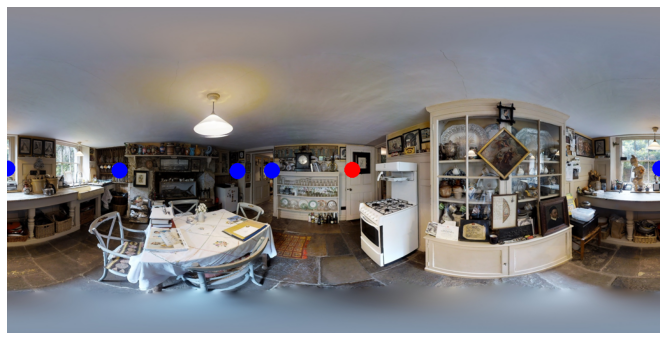} \vspace{2mm}\\
  \includegraphics[width=0.48\linewidth]{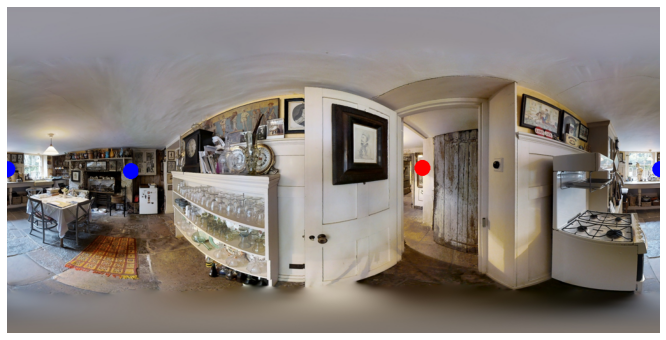}
  \includegraphics[width=0.48\linewidth]{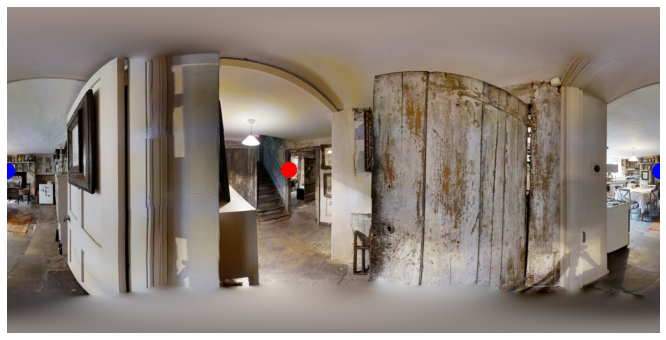} \vspace{2mm}\\
  \includegraphics[width=0.48\linewidth]{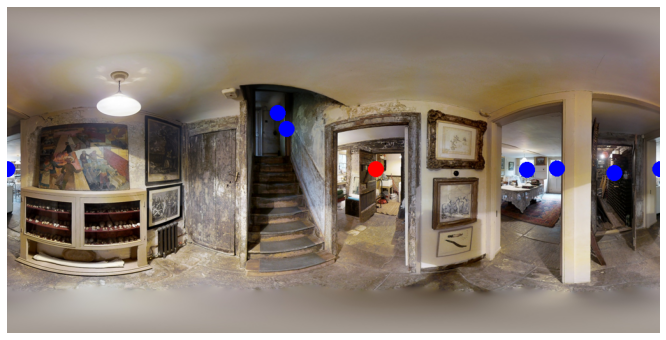}
  \includegraphics[width=0.48\linewidth]{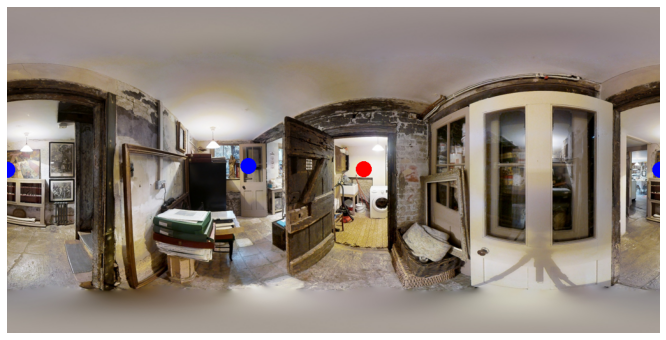} \vspace{2mm}\\
  \includegraphics[width=0.48\linewidth]{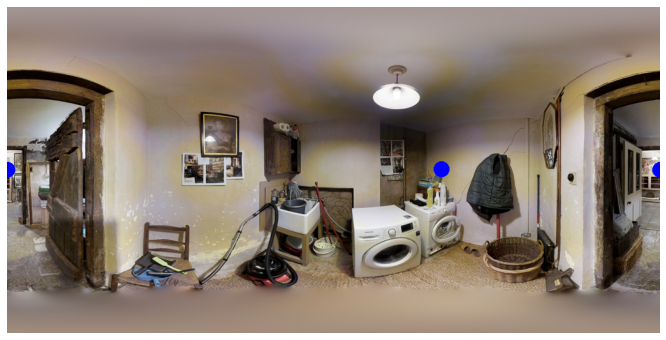}
  \caption{Inference example of \ModelName{} \textit{successfully} following an English instruction from the RxR Val-Unseen split through a sequence of panos. The panos are rotated so that the direction faced by the agent is the in center. \textcolor{blue}{Blue} dots indicate action candidates that the agent could move to. The selected action at each step is colored in \textcolor{red}{red}. }
  \label{fig:example1}
\end{figure*}

\begin{figure*}[h]
  \centering
  \textbf{Instruction:} You're starting in the corner of a living room. Turn around behind you and find the clock hanging on the wall in the hallway. Take two steps toward that. Turn right and go straight between the blue couch on your right and the kitchen counter on your left. You'll take three steps and stop at the first, right corner of the long dining table across from the kitchen. You should be looking through windows into the backyard in front of you. To the right, is an open patio, and to your left should be four framed, black and white pictures. You're done.
  \vspace{1mm}\\
  \includegraphics[width=0.48\linewidth]{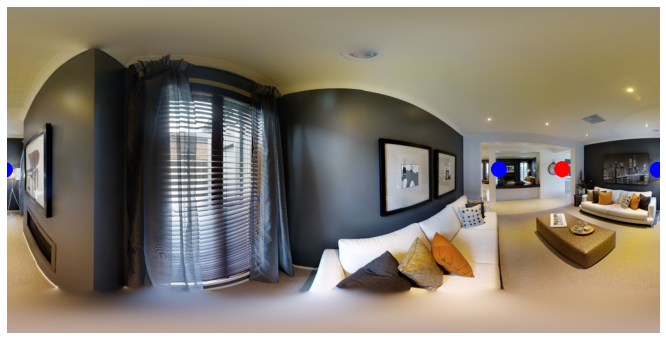}
  \includegraphics[width=0.48\linewidth]{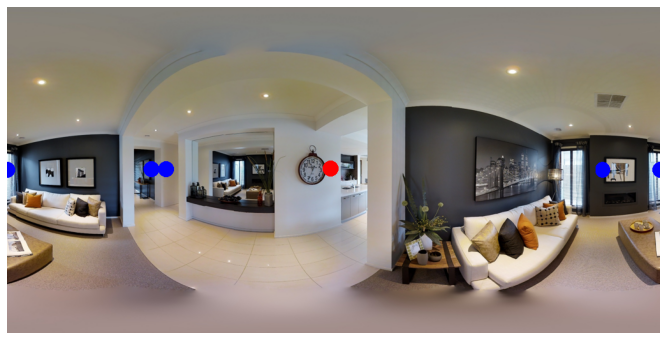} \vspace{2mm}\\
  \includegraphics[width=0.48\linewidth]{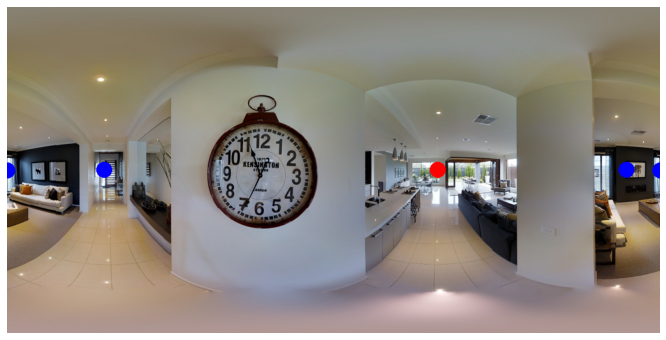}
  \includegraphics[width=0.48\linewidth]{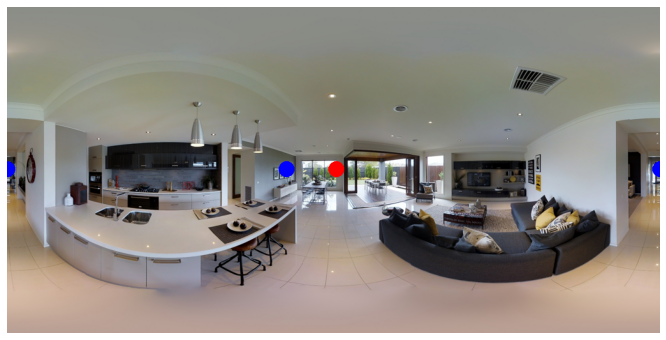} \vspace{2mm}\\
  \includegraphics[width=0.48\linewidth]{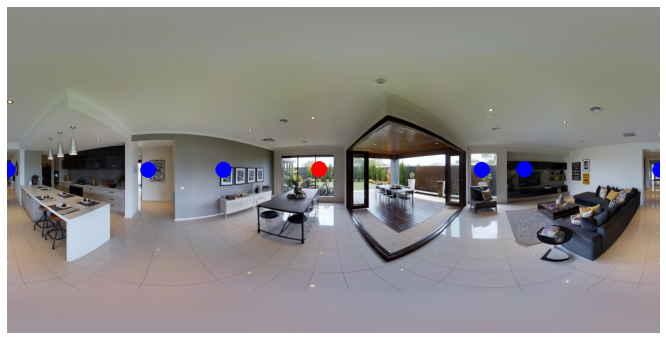}
  \includegraphics[width=0.48\linewidth]{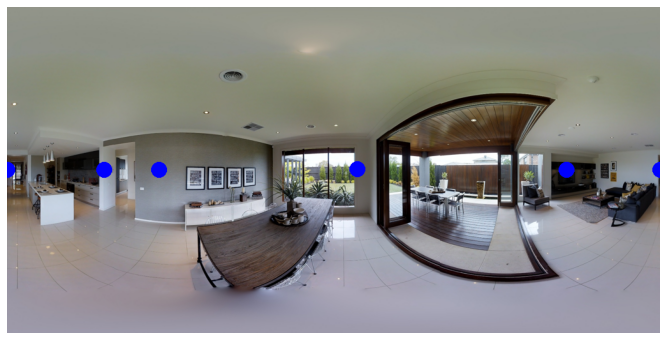}
  \caption{Inference example of \ModelName{} \textit{successfully} following an English instruction from the RxR Val-Unseen split through a sequence of panos.  The panos are rotated so that the direction faced by the agent is the in center. \textcolor{blue}{Blue} dots indicate action candidates that the agent could move to. The selected action at each step is colored in \textcolor{red}{red}. }
  \label{fig:example2}
\end{figure*}

\begin{figure*}[h]
  \centering
  \textbf{Instruction:} You are in the living room near a sofa chair and facing the open arch which is behind the sofa chair, move towards that arch, slightly turn right and walk down the narrow walkway, now move towards the closed wooden glass door which is in the front, turn left side and move towards the single chair which is on the right side, walk little forward from that single chair, now turn left side, and walk forward towards the small wooden cupboard in the front, slightly turn right side and enter the room, stand on the edge of the table which has book and a pen, that is your destination.
  \vspace{1mm}\\
  \includegraphics[width=0.48\linewidth]{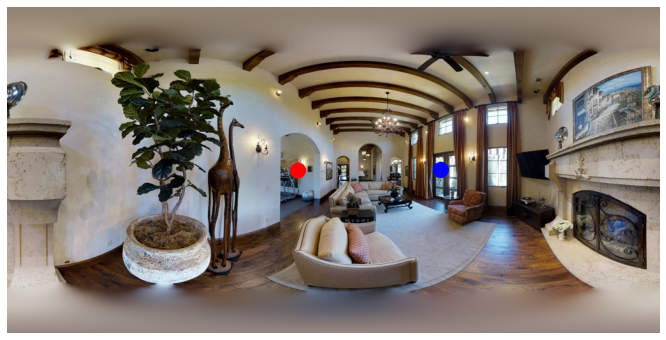}
  \includegraphics[width=0.48\linewidth]{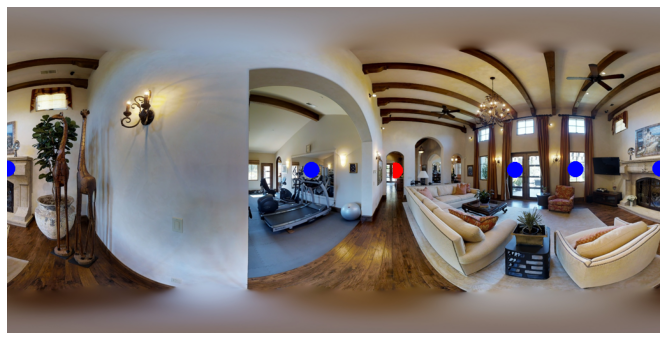} \vspace{2mm}\\
  \includegraphics[width=0.48\linewidth]{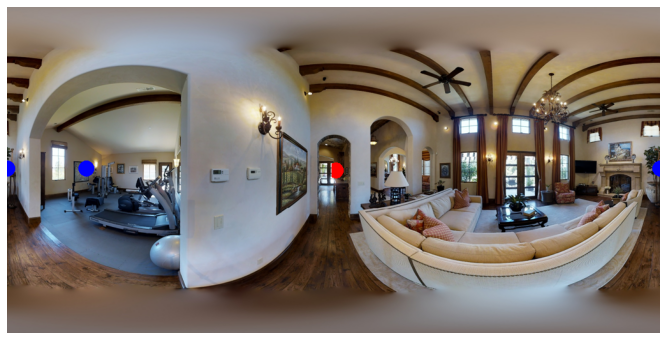}
  \includegraphics[width=0.48\linewidth]{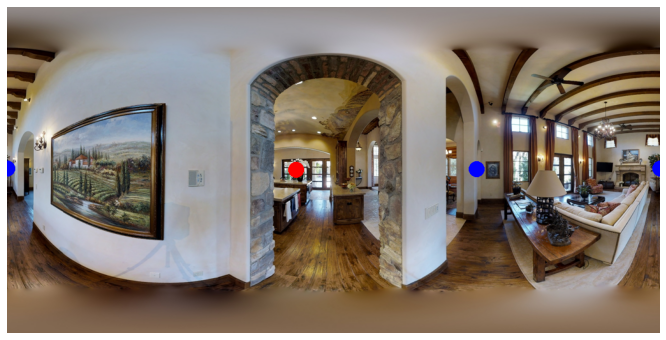} \vspace{2mm}\\
  \includegraphics[width=0.48\linewidth]{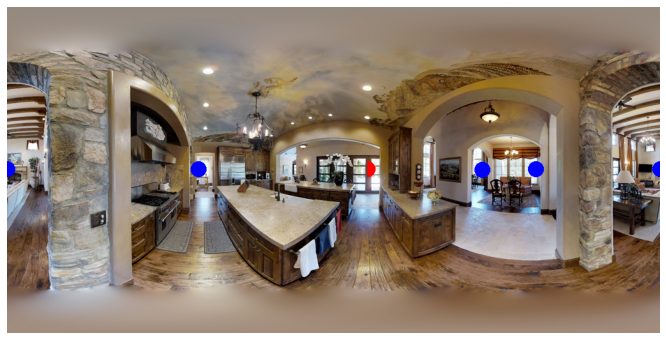}
  \includegraphics[width=0.48\linewidth]{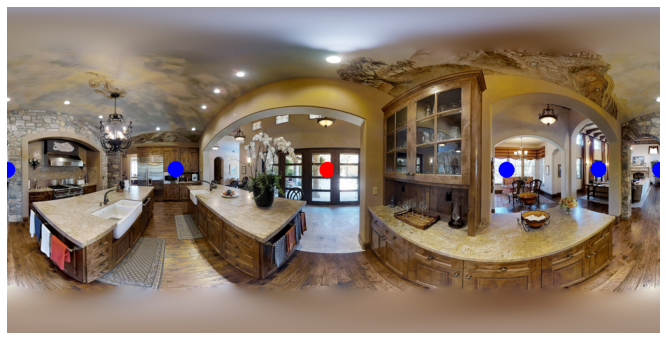} \vspace{2mm}\\
  \includegraphics[width=0.48\linewidth]{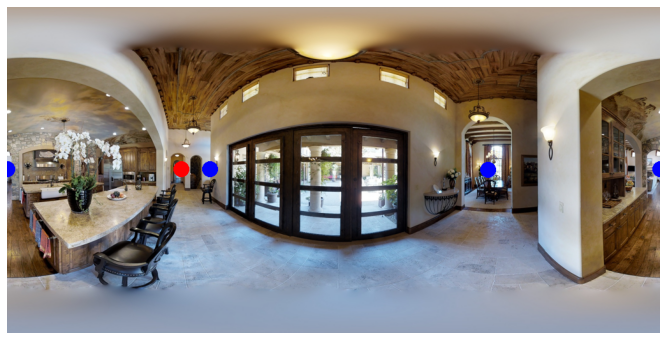}
  \includegraphics[width=0.48\linewidth]{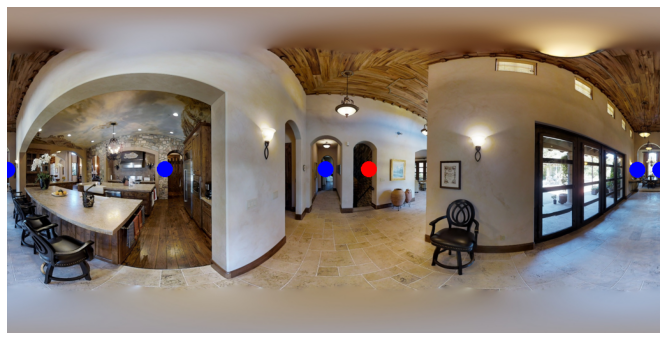}
  \caption{Inference example of \ModelName{} \textit{failing} to follow an English instruction from the RxR Val-Unseen split.
   The panos are rotated so that the direction faced by the agent is the in center. \textcolor{blue}{Blue} dots indicate action candidates that the agent could move to. The selected action at each step is colored in \textcolor{red}{red}. Here, \ModelName{} follows around 75\% of the instruction correctly, up until the instruction mentions the `single chair' (seen in the pano bottom right). However, at this point \ModelName{} makes an error and does not recover, failing to find the `table which has book and pen'. }
  \label{fig:example3}
\end{figure*}

Figures \ref{fig:example1} and \ref{fig:example2} provide examples \ModelName{} successfully following instructions from the RxR Val-Unseen split (i.e., in a previously unseen environment). In \figref{fig:example3} we include a failure case.

\end{document}